\documentclass[letterpaper]{article} % DO NOT CHANGE THIS
\usepackage{aaai25}  % DO NOT CHANGE THIS
\usepackage{times}  % DO NOT CHANGE THIS
\usepackage{helvet}  % DO NOT CHANGE THIS
\usepackage{courier}  % DO NOT CHANGE THIS
\usepackage[hyphens]{url}  % DO NOT CHANGE THIS
\usepackage{enumitem}
\usepackage{graphicx} % DO NOT CHANGE THIS
\usepackage{natbib}  % DO NOT CHANGE THIS AND DO NOT ADD ANY OPTIONS TO IT
\usepackage{caption} % DO NOT CHANGE THIS AND DO NOT ADD ANY OPTIONS TO IT
\usepackage{bm,amssymb,amsmath}
\usepackage{algorithm}
\usepackage{algorithmic}

\usepackage{color}
\usepackage{multirow}
\usepackage{newfloat}
\usepackage{listings}
\usepackage{booktabs}
\usepackage{float}
\usepackage{subfig}
\DeclareCaptionStyle{ruled}{labelfont=normalfont,labelsep=colon,strut=off} % DO NOT CHANGE THIS
\floatstyle{ruled}
\newfloat{listing}{tb}{lst}{}
\floatname{listing}{Listing}
\def\method{GSG}
\title{Geometry-Aware Spiking Graph Neural Network}
\author {
    Bowen Zhang\textsuperscript{\rm 1},
    Genan Dai\textsuperscript{\rm 1},
    Hu Huang\textsuperscript{\rm 2},
    Long Lan\textsuperscript{\rm 3}
}
\affiliations {
    \textsuperscript{\rm 1}Shenzhen Technology University\\
    \textsuperscript{\rm 2}University of Science and Technology of China\\
    \textsuperscript{\rm 3}National University of Defense Technology
}

\usepackage{bibentry}
\begin{document}

\maketitle

\begin{abstract}

Graph Neural Networks (GNNs) have demonstrated impressive capabilities in modeling graph-structured data, while Spiking Neural Networks (SNNs) offer high energy efficiency through sparse, event-driven computation. However, existing spiking GNNs predominantly operate in Euclidean space and rely on fixed geometric assumptions, limiting their capacity to model complex graph structures such as hierarchies and cycles. To overcome these limitations, we propose \method{}, a novel Geometry-Aware Spiking Graph Neural Network that unifies spike-based neural dynamics with adaptive representation learning on Riemannian manifolds. \method{} features three key components: a Riemannian Embedding Layer that projects node features into a pool of constant-curvature manifolds, capturing non-Euclidean structures; a Manifold Spiking Layer that models membrane potential evolution and spiking behavior in curved spaces via geometry-consistent neighbor aggregation and curvature-based attention; and a Manifold Learning Objective that enables instance-wise geometry adaptation through jointly optimized classification and link prediction losses defined over geodesic distances. All modules are trained using Riemannian SGD, eliminating the need for backpropagation through time. Extensive experiments on multiple benchmarks show that GSG achieves superior accuracy, robustness, and energy efficiency compared to both Euclidean SNNs and manifold-based GNNs, establishing a new paradigm for curvature-aware, energy-efficient graph learning.

\end{abstract}
\section{Introduction}

Graphs are ubiquitous non-Euclidean structures used to model complex relationships in real-world systems. Graph Neural Networks (GNNs)~\cite{hamilton2017inductive,velickovic2018graph,KipfW17}, built upon floating-point Artificial Neural Networks (ANNs), have achieved impressive success in learning expressive graph representations. However, their performance often comes at the cost of high computational and energy demands, particularly when scaling to large and complex graphs~\citep{zhu2022spiking,yin2024dynamic}. In parallel, Spiking Neural Networks (SNNs), inspired by the spike-based communication mechanism of biological neurons, offer event-driven and temporally sparse computation, demonstrating notable advantages in energy efficiency~\citep{maass1997networks,brette2007simulation}. Combining the structural modeling capability of GNNs with the energy efficiency of SNNs, spiking GNNs are rapidly gaining attention as a promising new paradigm in graph learning.

Despite recent advances in spiking GNNs, their representational expressiveness and adaptability remain fundamentally constrained by two key limitations. Prior efforts have primarily implemented spiking GNNs within Euclidean space~\citep{sarkar2011low}, inheriting the design of conventional artificial neural networks. Although this design is computationally convenient, it struggles to capture the intricate geometric structures present in many real-world graphs. For instance, hierarchical social networks~\citep{wang2022spiking} and molecular graphs with ring-like dependencies~\citep{gilmer2017neural,yang2019analyzing} naturally exhibit non-Euclidean characteristics, which cannot be accurately embedded into flat space without significant distortion. While hyperbolic and spherical manifolds have demonstrated greater suitability for preserving such structural properties~\citep{chami2019hyperbolic,xiong2022pseudo}, their integration into spiking GNNs has received little attention. Furthermore, most existing approaches assume a fixed geometric prior for the entire dataset, whether Euclidean or based on a specific manifold~\citep{chen2021fully,bachmann2020constant,xiong2022pseudo}. This global assumption overlooks the structural diversity across individual graphs and limits the model’s ability to adapt to varying local geometries. Although a few recent studies propose using mixed-curvature product spaces to improve representational flexibility~\citep{wang2023mixed,zhang2023data}, they adopt a shared geometric configuration across all samples. As a result, these methods are unable to adjust their geometric inductive biases in an instance-aware manner. These limitations highlight the need for a principled framework that can dynamically select the most appropriate geometry for each input graph, enabling spiking GNNs to more effectively align with the intrinsic structure of the data.

Addressing the aforementioned limitations introduces several significant technical challenges. First, there exists a fundamental modeling mismatch between the discrete, non-differentiable nature of spike signals in SNNs and the continuous, differentiable computations required for learning on Riemannian manifolds. Bridging this gap necessitates a principled mechanism that can map spike-driven dynamics into smooth geometric operations while preserving trainability. Second, there is currently no established formulation for describing spiking neural dynamics, such as membrane potential evolution, in non-Euclidean spaces. Defining such dynamics in a way that is both biologically plausible and geometrically consistent—ensuring, for example, that temporal evolution proceeds along geodesic paths—remains an open challenge. Third, existing methods lack the ability to adapt geometric representations on a per-instance basis. A unified framework capable of dynamically selecting or coordinating between multiple geometric spaces during training, in response to varying graph structures, must also ensure compatibility with the discrete nature of spike-based computation. These challenges collectively highlight the need for novel algorithmic and geometric tools that can unify spiking behavior with curvature-aware representation learning.

% \vspace{-1pt}
To address these challenges, we propose a \textbf{G}eometry-Aware \textbf{S}piking \textbf{G}raph Neural Network (\method{}). \method{} consists of three tightly integrated components. The Riemannian Embedding Layer mitigates the mismatch between flat Euclidean spaces and the intrinsic non-Euclidean structures in graph data by projecting Euclidean features onto a set of constant-curvature manifolds. This projection lays the foundation for curvature-aware representation and enables downstream modules to operate in geometry-aligned spaces. To bridge the gap between spiking dynamics and manifold geometry, the Manifold Spiking Graph Neural Network evolves membrane potentials in the tangent space and maps spike outputs to the manifold via exponential projection. This design ensures geometric consistency and differentiability through diffeomorphic mappings, effectively bridging discrete spikes and smooth manifolds. Lastly, the Manifold Learning Objective addresses the need for dynamic geometric adaptation by enabling geometry-consistent optimization across multiple manifolds. It supervises both node classification and link prediction using loss functions grounded in geodesic structures, allowing the model to learn instance-aware geometry while maintaining training stability without relying on recurrent backpropagation. These modules form a cohesive architecture that unifies geometry for expressive and efficient spiking graph representation learning.
Our contributions are summarized as follows:
% \vspace{-1pt}
\begin{itemize}[itemsep=2pt,topsep=0pt,parsep=0pt]
\item We present the framework to unify discrete spiking neural dynamics with continuous Riemannian geometry, introducing a novel paradigm of geometry-aware spiking representation learning.
\item We introduce \method{}, a unified architecture that integrates a manifold-consistent spiking mechanism, multi-space representation learning, and geodesic-aware optimization, enabling effective spike-based computation across mixed-curvature manifolds.
\item Extensive experiments on benchmarks demonstrate that \method{} achieves superior performance and energy efficiency compared to existing state-of-the-art models.
\end{itemize}
\section{Related work}

\paragraph{Spiking GNNs.}
Spiking Neural Networks (SNNs)~\citep{maass1997networks,brette2007simulation,cao2015spiking} draw inspiration from the firing patterns of biological neurons and achieve energy-efficient computation by transmitting information through sparse, binary spike signals in an event-driven manner. They offer significant energy advantages over traditional artificial neural networks, making them attractive for neuromorphic hardware and low-power applications~\citep{davies2018loihi,tavanaei2019deep}.
The integration of SNNs with graph structures has given rise to spiking GNNs, which adapt the spike-based computation paradigm to graph data~\citep{xu2021exploiting,zhu2022spiking}. Early designs focus on replacing continuous activations with spiking neurons in standard GNN architectures such as convolutional or attention-based models~\citep{li2023scaling,zhu2022spiking,sun2024spikegraphormer}. While promising, these approaches often rely on surrogate gradients and recurrent training schemes like Backpropagation Through Time (BPTT)~\citep{huh2018gradient}, which incur high computational latency.
More recently, researchers have sought to enhance both efficiency and representational power by introducing geometric awareness. The Manifold-valued Spiking GNN~\citep{sun2024spiking} introduces a novel spiking layer operating in Riemannian manifolds rather than Euclidean space, leveraging geodesic dynamics and diffeomorphisms to bridge spike trains and manifold representations. Crucially, it bypasses the BPTT bottleneck via a training paradigm called Differentiation via Manifold, enabling recurrence-free gradient propagation. However, current methods typically restrict representation learning to a single manifold, overlooking the diverse geometric structures in real-world graphs and thus limiting their expressivity.

\noindent\textbf{GNNs in Euclidean and Riemannian Spaces.}
Graph Neural Networks (GNNs) have become foundational tools for learning representations from graph-structured data~\citep{kipf2022semi,wu2019simplifying,velickovic2018graph}. Traditional GNNs typically operate in Euclidean space, which assumes flat geometry and offers computational simplicity. However, this assumption often leads to distortion when modeling real-world graphs with hierarchical, cyclical, or heterogeneous structures. To address these challenges, researchers have explored non-Euclidean spaces. Hyperbolic spaces, with negative curvature, are well-suited for hierarchical structures due to their exponential capacity, while spherical spaces, with positive curvature, better capture cyclic or angular patterns~\citep{coors2018spherenet}. These insights have motivated the development of Riemannian GNNs~\citep{chami2019hyperbolic,liu2019hyperbolic}, which generalize neural operations to curved manifolds such as hyperbolic, spherical, and their product spaces. Recent methods further propose mixed-curvature manifolds to better model structural diversity. However, most existing approaches adopt a fixed geometric space for all data, ignoring instance-level geometric variability. Our method addresses this limitation by dynamically selecting and combining multiple Riemannian spaces based on the structural properties of each input, resulting in more flexible and expressive graph representations.
\section{Preliminaries}
\paragraph{Problem Definition}
We address the problem of learning dynamic and geometry-aware representations over graph-structured data in an energy-efficient manner by integrating spiking neural networks (SNNs) with Riemannian manifolds. Given a graph $\mathcal{G} = (\mathcal{V}, \mathcal{E})$, where each node $v_i \in \mathcal{V}$ is associated with an input feature $\mathbf{x}_i \in \mathbb{R}^d$, the goal is to model temporally-evolving node states through spike-based dynamics. Each node emits a binary spike train $\{s_i^{(t)}\}_{t=1}^{T} \in \{0, 1\}^T$, determined by its manifold-valued membrane potential $u_i^{(t)} \in \mathcal{M}$, which is updated over time using neighborhood aggregation and geometry-aware operations.

To model structural heterogeneity (e.g., hierarchies or cycles), we embed node states in a mixed-curvature product space $\mathcal{M} = \mathcal{M}_1 \times \cdots \times \mathcal{M}_K$, where each $\mathcal{M}_k$ corresponds to a manifold with distinct curvature (e.g., hyperbolic, spherical, or Euclidean). The learning objective is to optimize a manifold-aware spiking graph network $f_\theta$, which maps temporal graph inputs into sparse and structured embeddings for downstream tasks. The overall training problem is formulated as:
$\min_{\theta} \mathcal{L}\big(f_\theta(\{ \mathbf{x}_i \}, \mathcal{G}, \mathcal{M})\big)$,
where $\mathcal{L}$ denotes a task-specific loss (e.g., cross-entropy for classification), and $f_\theta$ incorporates both the spiking neuron dynamics and the manifold geometry in its propagation and update rules.

\paragraph{Riemannian Geometry.}
To effectively model non-Euclidean structures in graph data, we embed node features into a constant-curvature Riemannian manifold $\mathcal{M}_\kappa^d$, where the curvature $\kappa \in \mathbb{R}$ determines the geometric characteristics of the space: spherical ($\kappa > 0$), flat Euclidean ($\kappa = 0$), or hyperbolic ($\kappa < 0$).

Given two distinct points $\mathbf{x}, \mathbf{y} \in \mathcal{M}\kappa^d$ and a tangent vector $\mathbf{t} \in \mathcal{T}_\mathbf{x} \mathcal{M}_\kappa^d$ (e.g., $\mathbf{t} \neq \mathbf{0} = [1/\sqrt{|\kappa|}, 0, \dots, 0]$), the transformation between the manifold and its tangent space is performed via the exponential and logarithmic maps:
\begin{equation}
    \exp^\kappa_\mathbf{x}(\cdot) : \mathcal{T}_\mathbf{x} \mathcal{M}\kappa^d \rightarrow \mathcal{M}\kappa^d, \quad \log^\kappa_\mathbf{x}(\cdot) : \mathcal{M}\kappa^d \rightarrow \mathcal{T}_\mathbf{x} \mathcal{M}_\kappa^d.\nonumber
\end{equation}

These maps enable learning in curved spaces by translating between nonlinear manifolds and linear tangent spaces. Their unified closed-form expressions are:
\begin{equation}
    \exp^\kappa_\mathbf{x}(\mathbf{t}) = \cos_\kappa\left( \sqrt{|\kappa|} \|\mathbf{t}\|_\kappa \right)\mathbf{x} + \sin_\kappa\left( \sqrt{|\kappa|} \|\mathbf{t}\|_\kappa \right) \frac{\mathbf{t}}{\sqrt{|\kappa|} \|\mathbf{t}\|_\kappa},\nonumber
\end{equation}
\begin{equation}
    \log^\kappa_\mathbf{x}(\mathbf{y}) = \frac{ \cos_\kappa^{-1} \left( \kappa \langle \mathbf{x}, \mathbf{y} \rangle_\kappa \right) }{ \sin_\kappa \left( \cos_\kappa^{-1} \left( \kappa \langle \mathbf{x}, \mathbf{y} \rangle_\kappa \right) \right)} \left( \mathbf{y} - \kappa \langle \mathbf{x}, \mathbf{y} \rangle_\kappa \mathbf{x} \right),\nonumber
\end{equation}
\begin{equation}
    d_\kappa(\mathbf{x}, \mathbf{y}) = \frac{1}{\sqrt{|\kappa|}} \cos_\kappa^{-1} \left( |\kappa| \langle \mathbf{x}, \mathbf{y} \rangle_\kappa \right),\nonumber
\end{equation}
where the inner product $\langle \cdot, \cdot \rangle_\kappa$ is generalized according to the manifold’s metric, and the generalized trigonometric functions are defined by curvature-specific rules:
\begin{equation}
    \cos_\kappa(z) =
\begin{cases}
\cos(\sqrt{\kappa} \, z), & \kappa > 0 \quad \text{(Spherical)} \\
1, & \kappa = 0 \quad \text{(Euclidean)} \\
\cosh(\sqrt{-\kappa} \, z), & \kappa < 0 \quad \text{(Hyperbolic)}
\end{cases},\nonumber
\end{equation}
\begin{equation}
    \sin_\kappa(z) =
\begin{cases}
\frac{1}{\sqrt{\kappa}} \sin(\sqrt{\kappa} \, z), & \kappa > 0 \\
z, & \kappa = 0 \\
\frac{1}{\sqrt{-\kappa}} \sinh(\sqrt{-\kappa} \, z), & \kappa < 0
\end{cases}.\nonumber
\end{equation}

% This unified formulation allows geometric computations to be consistently performed across different curvature regimes, laying the foundation for geometric learning in mixed-curvature spacetimes and enabling flexible integration with spiking graph neural models or temporal reasoning modules.

\paragraph{Spiking Neural Networks.}
Unlike conventional artificial neural networks that process continuous-valued activations, SNNs encode information through temporally distributed binary spikes. Each neuron integrates incoming spikes into a membrane potential over time and emits a spike once this potential surpasses a predefined threshold. The dynamics of first-order SNNs can be expressed as:
\begin{equation}
\begin{aligned}
u_{\tau+1,i} = \lambda ( u_{\tau,i} &- V_{th} \cdot s_{\tau,i} ) + \sum\nolimits_j w_{ij} s_{\tau,j} + b, \\
s_{\tau+1,i} &= \mathbb{H}(u_{\tau+1,i} - V_{th}), \nonumber
\end{aligned}
\end{equation}
where $\mathbb{H}(\cdot)$ denotes the Heaviside step function, representing the non-differentiable spike generation mechanism. Here, $s_{\tau,i} \in \{0,1\}$ indicates the binary spike state of neuron $i$ at time step $\tau$, $\lambda$ is the membrane decay constant, and $w_{ij}$, b are the synaptic weights and bias, respectively.

\section{Methodology}

% In this section, we detail the proposed GeoSNN framework. Our approach is designed to be general, applicable to any geodesically complete Riemannian manifold (e.g., hyperbolic or hyperspherical spaces) and their Cartesian products.

\subsection{The Overall Framework}
\begin{figure}[t!]
  \centering
  \includegraphics[width=0.48\textwidth]{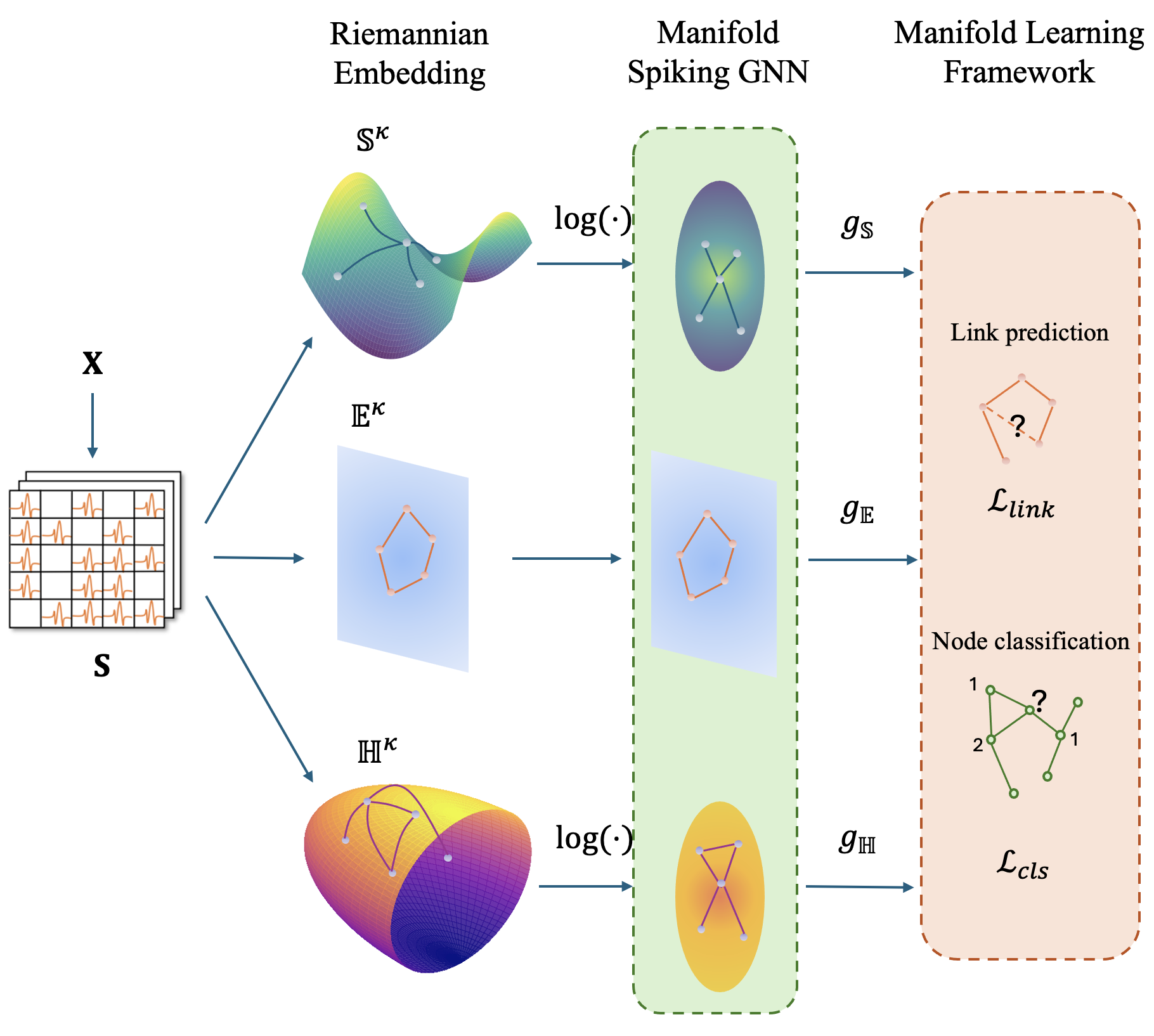} 
\caption{Overview of the proposed \method{}. \method{} consists of a Riemannian embedding layer that maps Euclidean node features to mixed-curvature manifolds for capturing non-Euclidean graph structures, a manifold spiking layer that integrates spike-based neural dynamics with curvature-aware message passing through geometric attention and nonlinearity, and a manifold learning objective that enables instance-wise geometry adaptation by jointly optimizing classification and link prediction using geodesic-based losses.}
  \label{framework}
  % \vspace{-0.4cm}
\end{figure}

This work explores the problem of geometry-aware spiking representation learning and introduces a novel framework, \method{}, as illustrated in Figure~\ref{framework}. \method{} consists of three core modules: (1) \textbf{Riemannian Embedding Layer.} To capture the intrinsic geometric structures of graph data, this module projects Euclidean node features onto constant-curvature manifolds via exponential mapping. The learned manifold embeddings enable curvature-aware spatial reasoning and provide a unified representation space for downstream processing; (2) \textbf{Manifold Spiking Graph Neural Network.} To integrate biologically inspired dynamics with geometric learning, we design a spiking GNN layer that evolves membrane potentials in the tangent space and maps spiking activations onto the manifold. A curvature-aware attention mechanism is employed to guide neighbor aggregation, while a Riemannian nonlinearity further enhances the expressive capacity of manifold-based spike representations; (3) \textbf{Manifold Learning Objective.} To support both node classification and link prediction, we formulate geometry-consistent loss functions that operate directly on manifold representations. A ranking-based link prediction loss leverages geodesic distances across multiple manifolds, while node classification is optimized using cross-entropy loss on log-mapped features. All parameters are optimized via Riemannian SGD to ensure geometry-preserving updates.

% \begin{figure}[h!]
%     \centering
%     % Placeholder for the actual figure.
%     \framebox[0.8\textwidth]{
%         \parbox{0.75\textwidth}{\centering \vspace{5cm} \large Figure Placeholder \vspace{5cm}}
%     }
%     \caption{Overall architecture of GeoSNN. An input data point $p$ is fed into the DCS module, which sparsely selects $K$ spaces from a geometric pool (e.g., selecting $\mathbb{P}$ and $\mathbb{E}$ from $\{\mathbb{P}, \mathbb{D}, \mathbb{E}\}$). The data is then processed through a stack of MNCM layers on the resulting Cartesian product manifold $\mathcal{M}_p = \mathbb{P} \times \mathbb{E}$. Each MNCM layer updates both the manifold representation and the spike trains. The entire model is optimized end-to-end via the MSOE.}
%     \label{fig:framework}
% \end{figure}
\subsection{Riemannian Embedding Layer}

To enable manifold-aware spatiotemporal reasoning, we project Euclidean node features into a Riemannian manifold with curvature $\kappa$. Let $\mathbf{x}_i^{\mathbb{E}} \in \mathbb{R}^d$ denote the initial embedding of node $n_i$. We first embed it into the tangent space $\mathcal{T}_{\mathbf{o}}\mathcal{M}_\kappa^d$ at a reference origin $\mathbf{o} = [\frac{1}{\sqrt{|\kappa|}}, 0, \dots, 0]^\top$ by forming:
$
\mathbf{v}_i = [0, \mathbf{x}_i^{\mathbb{E}}]^\top.
$
We then perform an exponential mapping $\exp^\kappa$ to place the node on the target curved manifold:
\begin{equation}
\label{exp}
    \mathbf{x}_i^{\mathcal{M}} \!=\! \exp^\kappa_{\mathbf{o}}(\mathbf{v}_i) \!=\! \left[\!\!
\begin{array}{c}
\frac{\cos_\kappa(\sqrt{|\kappa|} || \mathbf{x}_i^{\mathbb{E}} ||_2)}{\sqrt{|\kappa|}}\!,\! 
  \frac{\sin_\kappa(\sqrt{|\kappa|} || \mathbf{x}_i^{\mathbb{E}} ||_2)\mathbf{x}_i^{\mathbb{E}} }{\sqrt{|\kappa|} || \mathbf{x}_i^{\mathbb{E}} ||_2}
\end{array}
\!\right],
\end{equation}
where $\cos_\kappa$ and $\sin_\kappa$ are curvature-aware trigonometric functions defined as:
$$
\cos_\kappa(x) \!=\!
\begin{cases}
\cos(x) & \kappa > 0, \\
1 & \kappa = 0, \\
\cosh(x) & \kappa < 0,
\end{cases}
\;
\sin_\kappa(x) \!=\!
\begin{cases}
\sin(x) & \kappa > 0, \\
x & \kappa = 0, \\
\sinh(x) & \kappa < 0.
\end{cases}
$$

To support information propagation and neighborhood interaction, we also define the log map to pull points from the manifold back into the tangent space:
$$
\log^\kappa_{\mathbf{o}}(\mathbf{x}_i^{\mathcal{M}}) \!=\!
\left[
0,
\frac{\left( \mathbf{x}_i^{\mathcal{M}} - \cos_\kappa(d_\kappa(\mathbf{o}, \mathbf{x}_i^{\mathcal{M}}))\cdot \mathbf{o}  \right) d_\kappa(\mathbf{o}, \mathbf{x}_i^{\mathcal{M}})}{\sin_\kappa(d_\kappa(\mathbf{o}, \mathbf{x}_i^{\mathcal{M}}))}  
\right],
$$
where $d_\kappa(\cdot,\cdot)$ denotes the geodesic distance between points on $\mathcal{M}_\kappa^d$. The log mapping measures displacements in locally flat geometry, facilitating operations such as neighbor aggregation and attention in a mathematically consistent manner.

\subsection{Manifold Spiking GNN}
This module addresses the core incompatibility between discrete-time spiking dynamics and continuous curvature-aware representation learning in non-Euclidean spaces. Existing spiking GNNs are largely restricted to Euclidean geometry and rely on surrogate gradients and recurrence-based training (e.g., BPTT), which incur high computational cost and fail to capture intrinsic geometric structures such as hierarchies and cycles. These constraints limit both expressiveness and adaptability on complex graph data. To overcome this, we propose a manifold-aware spiking GNN that jointly models membrane potential dynamics and geometric representations on curved spaces. It integrates curvature-aware neighbor aggregation for geometry-consistent message passing, Riemannian non-linear activation for enhanced expressivity, and a query-guided attention mechanism that adaptively weights neighbors in the tangent space. This design unifies biologically plausible spiking with geometric learning, enabling both efficient and expressive representation.

\paragraph{Spiking GNN Layer.} 
Spiking GNNs achieve energy-efficient computation by simulating neural firing with sparse spike trains, but their representational capacity is fundamentally limited by their reliance on Euclidean space, which fails to capture the complex geometry of real-world graphs such as hierarchies and cycles. Moreover, the discrete nature of spike signals poses challenges for integration with smooth manifold representations, hindering the model’s ability to learn expressive geometric features.

To address these limitations, we propose a Manifold Spiking GNN Layer that seamlessly integrates spiking dynamics with geometry-aware representation learning. Specifically, given a node feature $\mathbf{x}_i^\mathbb{E} \in \mathbb{R}^d$, we compute a spike probability via sigmoid activation and generate a binary spike vector $\mathbf{s}_i^\mathbb{E} \sim \text{Bernoulli}(\mathbf{x}_i^\mathbb{E})$, providing temporally sparse inputs without requiring surrogate gradients or time-based integration. These spike representations are then projected into different Riemannian manifolds with Eq.~\eqref{exp}, enabling the model to capture non-Euclidean relational patterns while preserving the energy efficiency of spiking computation. 
Specifically, for a manifold $\mathcal{M}\in \{\mathbb{S},\mathbb{E},\mathbb{H}\}$, we have:
\begin{equation}
\begin{aligned}
    \mathbf{s}_i^{l+1,\mathcal{M}} = \exp_{\mathbf{s}_i^{l,\mathcal{M}}}^\kappa \left(\sum_{j\in \mathcal{N}(i)}\alpha_{ij}^{l,\mathcal{M}} \log_{\mathbf{s}_i^{l,\mathcal{M}}}^\kappa\left(\mathbf{s}_j^{l,\mathcal{M}}\right) \right),\nonumber
\end{aligned}
\end{equation}
\begin{equation}
    \mathbf{s}_i^{l+1,\mathbb{E}}= \text{IFModel}\left(\{\mathbf{s}_i^{l,\mathbb{E}}[t]\}_{t=1,\cdots,T}\right),\nonumber
\end{equation}
\begin{equation}
    \mathbf{s}_i^{l+1,\mathcal{M}} =\exp_{\mathbf{o}^\mathcal{M}}^\kappa\left(\mathbf{s}_i^{l+1,\mathbb{E}}\right),\nonumber
\end{equation}
where $\mathbf{s}_i^{l,\mathcal{M}}$ denotes the spike representation of node $i$ at layer $l$ in the manifold space $\mathcal{M} \in \{\mathbb{E}, \mathbb{S}, \mathbb{H}\}$, and $\mathbb{E}$, $\mathbb{S}$, and $\mathbb{H}$ represent Euclidean, spherical, and hyperbolic manifolds, respectively. $\mathcal{N}(i)$ denotes the set of neighbors of node $i$. The coefficient $\alpha_{ij}^{l,\mathcal{M}}$ is the attention weight assigned to neighbor $j$ when aggregating information for node $i$ in manifold $\mathcal{M}$. The operators $\log_{\mathbf{s}_i^{l,\mathcal{M}}}^{\kappa}(\cdot)$ and $\exp_{\mathbf{s}_i^{l,\mathcal{M}}}^{\kappa}(\cdot)$ represent the logarithmic and exponential maps at point $\mathbf{s}_i^{l,\mathcal{M}}$ on a manifold with curvature $\kappa$, projecting points from the manifold to the tangent space and vice versa. $\text{IFModel}(\cdot)$ denotes the integrate-and-fire spiking model that converts a temporal spike sequence into a final spike output. Finally, $\mathbf{o}^\mathcal{M}$ is a reference origin point (often the pole or canonical center) on the manifold $\mathcal{M}$ used for projecting Euclidean outputs into the corresponding manifold space via the exponential map.

\paragraph{Non-linear Activation Layer.}
After aggregating spiking representations on the manifold, we apply a non-linear transformation using Riemannian geometric operators to enhance the model’s expressive power. Specifically, we project the output to the tangent space, apply a hyperbolic tangent activation and map it to the manifold via the exponential map:
\begin{equation}
\mathbf{s}_i^{l+1,\mathcal{M}} = \exp_{\mathbf{o}^\mathcal{M}}^\kappa \left( \tanh \left( \log_{\mathbf{o}^\mathcal{M}}^\kappa \left( \mathbf{s}_i^{l+1,\mathcal{M}} \right) \right) \right),\nonumber
\end{equation}
where $\mathbf{s}i^{l+1,\mathcal{M}}$ is the aggregated manifold representation at layer $l+1$, and $\tanh(\cdot)$ denotes the element-wise hyperbolic tangent activation. The logarithmic map $\log{\mathbf{o}^\mathcal{M}}$ projects features onto the tangent space at the origin $\mathbf{o}^\mathcal{M}$, while the exponential map $\exp{\mathbf{o}^\mathcal{M}}^\kappa(\cdot)$ maps them back to the manifold. This operation introduces non-linearity in a curvature-aware manner, enabling better modeling of hierarchical and geometrically structured spiking patterns.

\paragraph{Curvature-based Attention for Aggregation.}
To enhance the expressiveness of attention aggregation under manifold geometry, we introduce a query-guided attention mechanism tailored for non-Euclidean spaces. Unlike traditional attention schemes that rely solely on pairwise similarity in Euclidean space, our formulation incorporates a manifold-specific query vector $\mathbf{q}_i^{l,\mathcal{M}}$ to better capture the intrinsic geometric semantics of each node. Intuitively, $\mathbf{q}_i^{l,\mathcal{M}}$ serves as an anchor direction in the tangent space of node $i$, guiding how incoming messages from neighbors should be weighted during aggregation.

Concretely, we first map each neighbor node $j$’s representation $\mathbf{s}_j^{l,\mathcal{M}}$ into the tangent space at $\mathbf{s}_i^{l,\mathcal{M}}$ via the logarithmic map, resulting in $\log_{\mathbf{s}_i^{l,\mathcal{M}}}^\kappa(\mathbf{s}_j^{l,\mathcal{M}})$. To evaluate the geometric alignment between node $i$ and its neighbors, we compute the similarity between the mapped neighbor and the query vector $\mathbf{q}_i^{l,\mathcal{M}}$, which is defined as:
\begin{equation}
    \mathbf{q}_i^{l,\mathcal{M}} = \mathbf{W}q \cdot \log_{\mathbf{s}_i^{l,\mathcal{M}}}^\kappa(\mathbf{s}_i^{l,\mathcal{M}}) + \mathbf{b}_q,\nonumber
\end{equation}
where $\mathbf{W}_q$ and $\mathbf{b}_q$ are learnable parameters. This construction allows the query to be dynamically adapted to the node’s own geometric context.
The resulting attention weights are computed via softmax-normalized similarities:
\begin{equation}
    \alpha_{ij}^{l,\mathcal{M}} =
\frac{\exp \left( \gamma_{ij}^{l,\mathcal{M}} \cdot
\text{sim} \left(
\log_{\mathbf{s}_i^{l,\mathcal{M}}}^\kappa(\mathbf{s}_j^{l,\mathcal{M}}),
\mathbf{q}_i^{l,\mathcal{M}}
\right) \right)}
{\sum\limits_{k \in \mathcal{N}(i)}
\exp \left( \gamma_{ik}^{l,\mathcal{M}} \cdot
\text{sim} \left(
\log{\mathbf{s}_i^{l,\mathcal{M}}}^\kappa(\mathbf{s}_k^{l,\mathcal{M}}),
\mathbf{q}_i^{l,\mathcal{M}}
\right) \right)},\nonumber
\end{equation}
where $\text{sim}(\cdot, \cdot)$ denotes the similarity metric in the tangent space (e.g., dot product), and $\gamma_{ij}^{l,\mathcal{M}}$ is a curvature-aware scaling factor. This formulation enables the model to selectively attend to neighbors based on their geometric relevance with respect to the node’s orientation on the manifold, leading to more discriminative and curvature-sensitive representations.

\subsection{Manifold Learning Framework}
Our manifold learning framework supports two distinct tasks: link prediction and node classification. Each task optimizes its own loss function independently, leveraging the geometry-aware spiking representations learned across mixed-curvature manifolds.

\textbf{Link Prediction.}
We formulate the link prediction task as a pairwise ranking problem, where the model is encouraged to assign higher scores to positive edges than to negative ones. Specifically, for each anchor node $u$, we sample a positive node $v_1$ (i.e., a true neighbor of $u$) and a negative node $v_2$ (i.e., a randomly sampled non-neighbor of $u$). The learning objective is to enforce a margin $m$ between the predicted scores of positive and negative edges. The loss is defined as:
$$
\mathcal{L}_{\text{link}} = \sum_{(u, v_1, v_2) \in \Omega} \max(0, r(u, v_1) + m - r(u, v_2)),
$$
where $\Omega$ denotes the set of sampled triplets and $m$ is a margin hyperparameter. The score function $r(u, v)$ is defined as:
\begin{equation}
    r(u, v) = \log \left( \sum_{\mathcal{M}} g_{\mathcal{M}}(\mathbf{s}_u^{\mathcal{M}}) \cdot  r_{\mathcal{M}}(\mathbf{s}_u^{\mathcal{M}}, \mathbf{s}_v^{\mathcal{M}}) \right),\nonumber
\end{equation}
\begin{equation}
    r_{\mathcal{M}}(u, v) = -d^2_{\mathcal{M}}\left(\exp^\kappa_0(\mathbf{s}_u^{\mathcal{M}}), \exp^\kappa_0(\mathbf{s}_v^{\mathcal{M}})\right),\nonumber
\end{equation}
\begin{equation}
    g_{\mathcal{M}}(\mathbf{s}_u^{\mathcal{M}})=\frac{f(\mathbf{s}_u^\mathcal{M})}{\sum_{i\in\mathcal{M}}f(\mathbf{s}_u^i)},\nonumber
\end{equation}
where $\exp^\kappa_0(\mathbf{s}_u)$ and $\exp^\kappa_0(\mathbf{s}_v)$ are the spiking representations of $u$ and $v$ in manifold $\mathcal{M}\in\{\mathbb{S},\mathbb{E},\mathbb{H}\}$, and $d^2_{\mathcal{M}}(\cdot, \cdot)$ denotes the squared geodesic distance on that space. $f(\cdot): \mathbb{R}^d \to \mathbb{R}$ is a neural mapping function that transforms node features into a scalar signal guiding the gating module.
$g_i(\mathbf{s}_u)$ is the gating coefficient produced from the features $\mathbf{s}_u^\mathcal{M})$, indicating the importance of the $i$-th space for node $u$. 
By integrating geometry-aware distances across multiple latent spaces, the model captures complex structural patterns and adaptively weighs their relevance, leading to more expressive and robust link predictions.

\textbf{Node Classification.}
For node classification, a labeled subset of nodes $v \in \mathcal{V}$ is used to supervise the learning of semantic representations. The model predicts class probabilities based on the final-layer manifold embeddings $\mathbf{s}_v^{\mathcal{M}}$, and the cross-entropy loss is defined as:
\begin{equation}
    \mathcal{L}_{\text{cls}} = - \sum_{v \in \mathcal{V}} \sum_{c=1}^{C} y_{v,c} \log \hat{y}_{v,c},
\end{equation}
where $y_{v,c} \in \{0, 1\}$ is the ground truth one-hot label for node $v$, and $\hat{y}_{v,c}$ is the predicted probability for class $c$ obtained via a softmax classifier. $C$ denotes the number of target classes. The predicted class probabilities $\hat{y}_{v,c}$ are obtained by applying a softmax function over the class logits $\mathbf{s}_{v}=\text{Softmax}\left(\log_{\mathbf{s}_{v}^{\mathcal{M}}}(\mathbf{s}_{v}^{\mathcal{M}})\right)$.

\begin{table*}[t]
\centering
\small
% \resizebox{\textwidth}{!}{
\begin{tabular}{l|cc|cc|cc|cc}
\toprule
\multirow{2}{*}{Method} & \multicolumn{2}{c|}{Computers} & \multicolumn{2}{c|}{Photo} & \multicolumn{2}{c|}{CS} & \multicolumn{2}{c}{Physics} \\
 & NC & LP & NC & LP & NC & LP & NC & LP \\
\midrule
% \multicolumn{9}{c}{\textit{ANN-E}} \\
GCN & 83.55{\scriptsize$\pm$0.71} & 92.07{\scriptsize$\pm$0.40} & 86.01{\scriptsize$\pm$0.20} & 88.84{\scriptsize$\pm$0.39} & 91.68{\scriptsize$\pm$0.84} & 93.68{\scriptsize$\pm$0.84} & 95.03{\scriptsize$\pm$0.19} & 93.46{\scriptsize$\pm$0.39} \\
GAT & 86.82{\scriptsize$\pm$0.04} & 91.91{\scriptsize$\pm$1.08} & 86.68{\scriptsize$\pm$1.32} & 88.45{\scriptsize$\pm$0.07} & 91.74{\scriptsize$\pm$0.22} & 94.06{\scriptsize$\pm$0.70} & 95.11{\scriptsize$\pm$0.29} & 93.44{\scriptsize$\pm$0.70} \\
SGC & 82.17{\scriptsize$\pm$1.25} & 90.46{\scriptsize$\pm$0.80} & 87.91{\scriptsize$\pm$0.65} & 89.84{\scriptsize$\pm$0.40} & 92.09{\scriptsize$\pm$0.05} & \textbf{95.94}{\scriptsize$\pm$0.43} & 94.77{\scriptsize$\pm$0.32} & 95.93{\scriptsize$\pm$0.70} \\
SAGE & 81.69{\scriptsize$\pm$0.86} & 90.56{\scriptsize$\pm$0.48} & 89.41{\scriptsize$\pm$1.28} & 89.86{\scriptsize$\pm$0.90} & \underline{92.71}{\scriptsize$\pm$0.73} & 95.22{\scriptsize$\pm$0.14} & \underline{95.62}{\scriptsize$\pm$0.26} & {95.75}{\scriptsize$\pm$0.37} \\
\midrule
% \multicolumn{9}{c}{\textit{ANN-R}} \\
HGCN & 88.71{\scriptsize$\pm$0.24} & 96.88{\scriptsize$\pm$0.53} & 89.18{\scriptsize$\pm$0.50} & 94.54{\scriptsize$\pm$0.20} & 90.72{\scriptsize$\pm$0.16} & 93.02{\scriptsize$\pm$0.26} & 94.46{\scriptsize$\pm$0.20} & 94.10{\scriptsize$\pm$0.64} \\
$\kappa$-GCN & 89.20{\scriptsize$\pm$0.50} & 95.30{\scriptsize$\pm$0.24} & 92.22{\scriptsize$\pm$0.62} & 94.89{\scriptsize$\pm$0.15} & 91.98{\scriptsize$\pm$0.16} & 94.86{\scriptsize$\pm$0.18} & {95.85}{\scriptsize$\pm$0.20} & 94.58{\scriptsize$\pm$0.22} \\
Q-GCN & 85.94{\scriptsize$\pm$0.93} & \underline{96.98}{\scriptsize$\pm$0.05} & 92.50{\scriptsize$\pm$0.95} & \underline{97.47}{\scriptsize$\pm$0.03} & 91.18{\scriptsize$\pm$0.28} & 93.39{\scriptsize$\pm$0.20} & 94.84{\scriptsize$\pm$0.25} & OOM \\
HyboNet & 86.29{\scriptsize$\pm$2.30} & 96.80{\scriptsize$\pm$0.05} & {92.67}{\scriptsize$\pm$0.09} & {97.70}{\scriptsize$\pm$0.07} & {92.34}{\scriptsize$\pm$0.03} & 95.65{\scriptsize$\pm$0.26} & 95.56{\scriptsize$\pm$0.18} & \textbf{98.46}{\scriptsize$\pm$0.49} \\
\midrule
% \multicolumn{9}{c}{\textit{SNN-E}} \\
SpikeNet & 88.00{\scriptsize$\pm$0.70} & - & 92.90{\scriptsize$\pm$0.10} & - & 92.15{\scriptsize$\pm$0.18} & - & 92.66{\scriptsize$\pm$0.30} & - \\
SpikeGCN & 86.90{\scriptsize$\pm$0.30} & 91.12{\scriptsize$\pm$1.79} & 92.60{\scriptsize$\pm$0.70} & 93.84{\scriptsize$\pm$0.03} & 90.86{\scriptsize$\pm$0.87} & {95.07}{\scriptsize$\pm$1.22} & 94.53{\scriptsize$\pm$0.18} & 92.88{\scriptsize$\pm$0.80} \\
SpikeGCL & 89.04{\scriptsize$\pm$0.89} & {92.72}{\scriptsize$\pm$0.03} & {92.50}{\scriptsize$\pm$0.17} & {95.58}{\scriptsize$\pm$0.11} & 91.77{\scriptsize$\pm$0.26} & {95.13}{\scriptsize$\pm$0.24} & 95.21{\scriptsize$\pm$0.10} & {94.15}{\scriptsize$\pm$0.35} \\
SpikeGT & 81.00{\scriptsize$\pm$1.06} & - & 90.66{\scriptsize$\pm$0.38} & - & 91.86{\scriptsize$\pm$0.41} & - & 94.38{\scriptsize$\pm$1.57} & - \\
MSG & \underline{89.27}{\scriptsize$\pm$0.19} & 94.65{\scriptsize$\pm$0.73} &  \underline{93.11}{\scriptsize$\pm$0.11} & 96.75{\scriptsize$\pm$0.18} & {92.65}{\scriptsize$\pm$0.04} & {95.19}{\scriptsize$\pm$0.15} & \underline{95.93}{\scriptsize$\pm$0.07} & 93.43{\scriptsize$\pm$0.16} \\
\midrule
\method{} & \textbf{90.11}{\scriptsize$\pm$0.27} & \textbf{97.27}{\scriptsize$\pm$0.92} &  \textbf{93.62}{\scriptsize$\pm$0.30} & \textbf{97.75}{\scriptsize$\pm$0.37} & \textbf{93.01}{\scriptsize$\pm$0.27} & \underline{95.82}{\scriptsize$\pm$0.86} & \textbf{96.27}{\scriptsize$\pm$0.13} & \underline{97.46}{\scriptsize$\pm$0.73}\\
\bottomrule
\end{tabular}
\caption{Node Classification (NC) accuracy (\%) and Link Prediction (LP) AUC (\%) on four datasets. The best results are \textbf{boldfaced}, and the runner-ups are \underline{underlined}.}
% }
% \vspace{-0.4cm}
\label{tab:main_results}
\end{table*}

We optimize both tasks using Riemannian SGD~\cite{bonnabel2013stochastic}, which performs gradient updates that respect the underlying geometry of the manifold space. Unlike conventional Euclidean optimization, Riemannian SGD ensures that each update step remains on the manifold by projecting the gradients onto the tangent space and retracting them back to the curved space after each iteration. This enables more stable and geometry-preserving training dynamics. All model parameters are updated jointly, and the classification loss is minimized with task-specific learning rates and curvature-aware optimization trajectories.
\section{Experiments}

\subsection{Experimental Settings}
\textbf{Datasets.}
We evaluate the proposed method \method{} on four widely adopted benchmark datasets, spanning both co-purchase and co-authorship domains. Specifically, the Computers and Photo datasets~\citep{shchur2018pitfalls} are derived from Amazon co-purchase relationships, while the CS and Physics datasets~\citep{shchur2018pitfalls} represent academic co-authorship networks. These datasets capture diverse graph characteristics in terms of domain, connectivity, and node label semantics, providing a comprehensive testbed for validating the effectiveness of our model across varying structural and relational settings. 
% Additional statistics and preprocessing details are provided in Appendix~\ref{appendix:data}.

\noindent\textbf{Baselines.}
To benchmark the performance of our proposed \method{}, we compare it with twelve strong baselines that fall into three main categories. The first category consists of classical Euclidean graph neural networks, including GCN~\citep{kipf2022semi}, GAT~\citep{velivckovic2018graph}, GraphSAGE~\citep{hamilton2017inductive}, and SGC~\citep{wu2019simplifying}, which operate in flat geometric spaces using standard message passing. The second category includes Riemannian GNNs such as HGCN~\citep{chami2019hyperbolic} and HyboNet~\citep{chen2021fully} for hyperbolic embeddings, $\kappa$-GCN~\citep{bachmann2020constant} for constant curvature spaces, and Q-GCN~\citep{xiong2022pseudo} which exploits quotient space geometry. The third category comprises existing spiking-based graph models including SpikeNet~\citep{li2023scaling}, SpikeGCN~\citep{zhu2022spiking}, SpikeGraphormer ~\citep{sun2024spikegraphormer} (denote as SpikeGT), and the recent SpikeGCL~\citep{li2023graph}. Although some of these models were proposed for dynamic graph scenarios, we adapt them to static graphs to ensure fair comparison~\citep{li2023graph}. Notably, spiking GNNs have not been explored in the context of Riemannian geometry, our work aims to fill this gap by integrating manifold learning with biologically inspired spiking representations.

\noindent\textbf{Implementation Details.}
The proposed \method{} is implemented to operate on constant-curvature manifolds and is instantiated using a hyperbolic geometry with curvature-aware operations such as exponential mapping and metric-adaptive gradients. The embedding dimension in each space is fixed to 32. We employ an integrate-and-fire (IF) spiking neuron model with simulation time steps $T \in \{5, 15\}$, and the model is optimized using Riemannian SGD to ensure updates remain on the manifold. The learning rate is selected via grid search from $\{0.001, 0.003\}$ for node classification, and the dropout rate is tuned from $\{0.1, 0.3, 0.5\}$ for link prediction. The geometric step size is set to 0.1 throughout. 

\subsection{Performance Comparison}

Table~\ref{tab:main_results} reports the results of node classification (NC) and link prediction (LP) across four benchmark datasets, comparing the proposed GSG with twelve strong baselines, including Euclidean GNNs, manifold-based GNNs, and spiking GNNs. Our method consistently achieves state-of-the-art performance on both tasks, validating its effectiveness in addressing the core challenges identified earlier.
(1) Compared to Euclidean GNNs (e.g., GCN, GAT, SAGE), GSG exhibits significant improvements, particularly on datasets Physics and Photo. This underscores the importance of operating in non-Euclidean spaces and demonstrates that our manifold-aware spiking architecture successfully captures intrinsic structural patterns that flat-space models fail to preserve—tackling the first challenge of limited expressiveness due to Euclidean assumptions.
(2) While existing manifold-based GNNs such as $\kappa$-GCN and HyboNet also leverage curvature-aware representations, they adopt fixed geometric priors. In contrast, GSG introduces instance-wise geometry adaptation via its manifold learning objective, achieving superior performance (e.g., 96.27\% NC on Physics and 97.75\% LP on Photo), thereby addressing the third challenge regarding the lack of geometric flexibility across samples. This dynamic adaptation enables GSG to outperform even strong baselines like HyboNet and Q-GCN on most tasks.
(3) In comparison with prior spiking GNNs (e.g., SpikeGCN, SpikeGCL, SpikeGT), which are limited to Euclidean space and rely on surrogate gradients, GSG integrates biologically plausible spiking dynamics directly into curved manifolds. This unifies discrete spike behavior with continuous geometry, effectively resolving the challenge of incompatibility between spike-based computation and smooth manifold optimization. The performance gains in both tasks across all datasets demonstrate the expressiveness and stability brought by this principled integration.

\begin{table}[t]
\centering
\small
\setlength{\tabcolsep}{2.5pt}
\label{tab:ablation_geometry}
% \resizebox{0.48\textwidth}{!}{
\begin{tabular}{lcccc}
\toprule
\textbf{Geometry} & \textbf{Computers} & \textbf{Photo} & \textbf{CS} & \textbf{Physics} \\
\midrule
$\mathbb{H}^{32}$ & {89.27}{\scriptsize$\pm$0.19} & {93.11}{\scriptsize$\pm$0.11} & 92.65{\scriptsize$\pm$0.04} & {95.93}{\scriptsize$\pm$0.07} \\
$\mathbb{S}^{32}$ & 87.84{\scriptsize$\pm$0.77} & 92.03{\scriptsize$\pm$0.79} & 92.72{\scriptsize$\pm$0.06} & 95.85{\scriptsize$\pm$0.02} \\
$\mathbb{E}^{32}$ & 88.94{\scriptsize$\pm$0.24} & 92.93{\scriptsize$\pm$0.21} & {92.82}{\scriptsize$\pm$0.04} & 95.81{\scriptsize$\pm$0.04} \\
\midrule
$\mathbb{H}^{16} \times \mathbb{H}^{16}$ & 89.18{\scriptsize$\pm$0.25} & 92.06{\scriptsize$\pm$0.14} & 92.67{\scriptsize$\pm$0.10} & 95.90{\scriptsize$\pm$0.04} \\
$\mathbb{H}^{16} \times \mathbb{S}^{16}$ & 88.00{\scriptsize$\pm$1.05} & 91.97{\scriptsize$\pm$0.08} & 92.33{\scriptsize$\pm$0.21} & 95.73{\scriptsize$\pm$0.11} \\
$\mathbb{S}^{16} \times \mathbb{S}^{16}$ & 82.49{\scriptsize$\pm$1.18} & 92.31{\scriptsize$\pm$0.45} & 92.18{\scriptsize$\pm$0.21} & 95.81{\scriptsize$\pm$0.10} \\
\midrule
$\mathbb{S}^{8}\times \mathbb{S}^{8} \times \mathbb{H}^{8}$ & {89.69}{\scriptsize$\pm$0.33} &  {93.07}{\scriptsize$\pm$0.34} & {92.47}{\scriptsize$\pm$0.19} & {96.08}{\scriptsize$\pm$0.20} \\
$\mathbb{S}^{16}\times \mathbb{S}^{8} \times \mathbb{H}^{4}$ & \textbf{90.13}{\scriptsize$\pm$0.15} &  {93.41}{\scriptsize$\pm$0.21} & {92.63}{\scriptsize$\pm$0.17} & {95.67}{\scriptsize$\pm$0.23} \\
$\mathbb{S}^{4}\times \mathbb{S}^{8} \times \mathbb{H}^{16}$ & {90.11}{\scriptsize$\pm$0.27} &  \textbf{93.62}{\scriptsize$\pm$0.30} & \textbf{93.01}{\scriptsize$\pm$0.27} & \textbf{96.27}{\scriptsize$\pm$0.13} \\
\bottomrule
\end{tabular}
\caption{Ablation study of geometric variants. Results of node classification in terms of ACC (\%).}
% }
% \vspace{-0.4cm}
\end{table}

\subsection{Ablation Study}

To further evaluate the influence of manifold selection on representation learning, we conduct ablation experiments using nine geometric variants of our model, where node embeddings are projected onto different constant-curvature manifolds or their product spaces. Specifically, we instantiate the model using individual manifolds, hyperbolic ($\mathbb{H}^{32}$), spherical ($\mathbb{S}^{32}$), and Euclidean ($\mathbb{E}^{32}$), as well as three product spaces combining different curvature types, with total embedding dimension kept constant.

As shown in Table 2, $\mathbb{H}^{32}$ achieves the best overall performance among the single-manifold variants, confirming the suitability of hyperbolic geometry for modeling hierarchical or tree-like structures commonly found in real-world graphs. Moreover, multi-space product variants, particularly $\mathbb{S}^4 \times \mathbb{S}^8 \times \mathbb{H}^{16}$, achieve the highest accuracy on most datasets, demonstrating that combining curvature types offers better flexibility and expressiveness. These findings validate the importance of instance-level geometry adaptation, and directly support our design goal of enabling the model to dynamically align with diverse graph topologies.

\subsection{Energy Cost Analysis}

To assess the computational efficiency of different models, we report the estimated inference-time energy cost and parameter footprint across all datasets. The energy cost of the graph models in terms of theoretical energy
consumption (mJ)~\citep{zhu2022spiking}. Each method is evaluated under uniform simulation settings to ensure fairness in comparison.

As shown in Table 3, spiking-based models consistently achieve lower energy consumption than traditional ANN-based GNNs, owing to their binary, event-driven computation. Among them, MSG exhibits the lowest energy footprint across most datasets, reflecting its design emphasis on minimal power usage. While our proposed GSG incurs moderately higher energy cost and parameter count than MSG, it delivers a clear performance advantage. This reflects a deliberate trade-off: by slightly sacrificing energy efficiency, GSG achieves superior task accuracy through enhanced geometric modeling and dynamic curvature adaptation. Such a balance makes GSG well-suited for scenarios where both efficiency and representation fidelity are critical.

\begin{table}[t]
\centering
\footnotesize
\label{tab:energy}
\setlength{\tabcolsep}{2.5pt}
% \resizebox{0.48\textwidth}{!}{
\begin{tabular}{l|cc|cc|cc}
\toprule
\multirow{2}{*}{Method} & \multicolumn{2}{c|}{Computers} & \multicolumn{2}{c|}{Photo} & \multicolumn{2}{c}{CS}  \\
 & \#(para.) & energy & \#(para.) & energy & \#(para.) & energy  \\
\midrule
% \multicolumn{9}{c}{\textit{ANN-E}} \\
GCN & 24.91 & 1.671 & 24.14 & 0.893 & 218.29 & 18.444  \\
GAT & 24.99 & 2.477 & 24.22 & 1.273 & 218.38 & 28.782 \\
SGC & \textbf{7.68} & {0.508} & \textbf{5.97} & {0.219} & \textbf{102.09} & {8.621}  \\
SAGE & 49.77 & 1.671 & 48.23 & 0.893 & 436.53 & 18.444  \\
\midrule
% \multicolumn{9}{c}{\textit{ANN-R}} \\
HGCN & 24.94 & 1.614 & 24.96 & 0.869 & 217.79 & 18.390 \\
$\kappa$-GCN & 25.89 & 1.647 & 25.12 & 0.889 & 218.24 & 18.440  \\
Q-GCN & 24.93 & 1.629 & 24.96 & 0.876 & 217.83 & 18.393 \\
HyboNet & 27.06 & 1.625 & 26.29 & 0.875 & 219.94 & 18.399  \\
\midrule
% \multicolumn{9}{c}{\textit{SNN-E}} \\
SpikeNet & 101.22 & \underline{0.070} & 98.07 & \textbf{0.040} & 438.51 & 0.218 \\
SpikingGCN & 38.40 & 0.105 & 29.84 & 0.046 & 510.65 & 1.871 \\
SpikeGCL & 59.26 & 0.121 & 57.85 & 0.067 & 445.69 & \underline{0.128} \\
SpikeGT & 77.07 & 1.090 & 74.46 & 0.584 & 365.28 & 6.985 \\
MSG  & \underline{26.95} & \textbf{0.047} & \underline{25.68} & \underline{0.043} & \underline{226.15} & \textbf{0.026}  \\
\midrule
\method{} &54.72 &0.096  &52.80 &0.079 &438.71 &0.051  \\
\bottomrule
\end{tabular}
\caption{Energy cost. The number of parameters at runtime (KB) and theoretical energy consumption (mJ) on Computers, Photo, and CS datasets. The best results are \textbf{boldfaced}, and the runner-ups are \underline{underlined}.}
% }
% \vspace{-0.4cm}
\end{table}

% Among all models, our proposed MSG achieves the best or second-best energy efficiency on most datasets. In particular, MSG consumes the least energy on Computers, CS, and Physics datasets, outperforming both Euclidean and Riemannian GNNs, as well as other spiking baselines. Despite using a moderate number of parameters, MSG maintains extremely low energy overhead, requiring at least 20$\times$ less energy than Riemannian GNNs while delivering comparable or superior accuracy. This highlights the effectiveness of learning manifold-aware spiking representations not only in performance but also in energy-aware scenarios.

\section{Conclusion}

We propose GSG, a Geometry-Aware Spiking Graph Neural Network that unifies spike-based neural computation with manifold-aware representation learning. To overcome the limitations of existing spiking GNNs in modeling complex graph structures, GSG integrates a Riemannian embedding layer to project node features into mixed-curvature spaces, a manifold spiking layer for curvature-consistent message passing with biologically plausible dynamics, and a geodesic-aware learning objective enabling instance-wise geometry adaptation. This unified design captures both topological complexity and temporal sparsity, offering a principled approach to structure-aware and energy-efficient graph learning. Extensive experiments show that GSG consistently outperforms Euclidean, manifold-based, and spiking baselines in both accuracy and robustness, while maintaining competitive energy efficiency. Ablation studies further validate the contribution of each geometric component. These results underscore the potential of GSG as a general-purpose framework for expressive and efficient spiking graph representation learning. In future work, we plan to extend GSG to dynamic graphs and neuromorphic hardware to explore its scalability and real-world applicability.

% In this paper, we extend the research domain of Spiking Neural Networks to Riemannian manifolds for the first time, proposing a novel graph neural network framework named GeoSNN. Through an innovative Manifold-Neuron Coupling Module, a data-driven Dynamic Curvature Selector, and an efficient Manifold-Sparse Optimization Engine, GeoSNN successfully combines the energy efficiency of SNNs with the powerful expressivity of Riemannian geometry. It not only adaptively selects the optimal geometric representation space for each data point but also achieves efficient and stable optimization through a novel, BPTT-free training paradigm. Experimental results on multiple benchmarks robustly demonstrate the superior performance and efficiency of our method. This work opens up new avenues for developing next-generation neural models that are simultaneously biologically inspired, structurally adaptive, and representationally flexible.

% \clearpage
\bibliography{aaai25}

% \clearpage
% \def\isChecklistMainFile{}
% \input{checklist/ReproducibilityChecklist}

% \appendix
% \input{code/6_appendix}

\end{document}